\documentclass[sigconf]{acmart}
\AtBeginDocument{%
  \providecommand\BibTeX{{%
    \normalfont B\kern-0.5em{\scshape i\kern-0.25em b}\kern-0.8em\TeX}}}

\acmConference[ICAIL 2021]{ICAIL '21: The 18th International Conference on Artificial Intelligence 
and Law (ICAIL 2021)}{June 21--25, 2021}{Online}
\acmBooktitle{ICAIL 2021: The 18th International Conference on Artificial Intelligence 
and Law,  June 21--25, 2021, Online}

\setcopyright{rightsretained}
\acmConference[COLIEE 2021]{COLIEE 2021 workshop: Competition on Legal
  Information Extraction/Entailment }{June 21, 2021}{Online}
\acmISBN{}
\acmDOI{}

\renewcommand\footnotetextcopyrightpermission[1]{}

\usepackage{CJKutf8} 
\usepackage{svg}

\usepackage[normalem]{ulem}
\useunder{\uline}{\ul}{}

\begin{document}


\title{ParaLaw Nets - Cross-lingual Sentence-level Pretraining\\ for Legal Text Processing}

\author{Ha-Thanh Nguyen}
\affiliation{%
  \institution{Japan Advanced Institute of Science and Technology}
  \city{Ishikawa}
  \country{Japan}
}

\author{Vu Tran}
\affiliation{%
  \institution{Japan Advanced Institute of Science and Technology}
  \city{Ishikawa}
  \country{Japan}
}

\author{Phuong Minh Nguyen}
\affiliation{%
  \institution{Japan Advanced Institute of Science and Technology}
  \city{Ishikawa}
  \country{Japan}
}

\author{Thi-Hai-Yen Vuong}
\affiliation{%
  \institution{University of Engineering and Technology, VNU}
  \city{Hanoi}
  \country{Vietnam}
}

\author{Quan Minh Bui} 
\affiliation{%
  \institution{Japan Advanced Institute of Science and Technology}
  \city{Ishikawa}
  \country{Japan}
}

\author{Chau Minh Nguyen} 
\affiliation{%
  \institution{Japan Advanced Institute of Science and Technology}
  \city{Ishikawa}
  \country{Japan}
}

\author{Binh Tran Dang}
\affiliation{%
  \institution{Japan Advanced Institute of Science and Technology}
  \city{Ishikawa}
  \country{Japan}
}

\author{Minh Le Nguyen}
\affiliation{%
  \institution{Japan Advanced Institute of Science and Technology}
  \city{Ishikawa}
  \country{Japan}
}


\author{Ken Satoh}
\affiliation{%
  \institution{National Institute of Informatics}
  \city{Tokyo}
  \country{Japan}
}

\renewcommand{\shortauthors}{Nguyen et al.}

\begin{abstract}
  Ambiguity is a characteristic of natural language, which makes expression ideas flexible. However, in a domain that requires accurate statements, it becomes a barrier.
  Specifically, a single word can have many meanings and multiple words can have the same meaning. 
  When translating a text into a foreign language, the translator needs to determine the exact meaning of each element in the original sentence to produce the correct translation sentence.
  From that observation, in this paper, we propose ParaLaw Nets, a pretrained model family using sentence-level cross-lingual information to reduce ambiguity and increase the performance in legal text processing.
  This approach achieved the best result in the Question Answering task of COLIEE-2021.
\end{abstract}

\ccsdesc[500]{Computing methodologies~Neural networks}
\ccsdesc[300]{Applied computing~Law}

\keywords{pretrained model, legal text processing, cross-lingual, sentence-level}


\maketitle

\section{Introduction}
\label{section:introduction}
Transformer~\cite{vaswani2017attention}, the architecture using encoders and decoders with the attention mechanism has become the best practice in many problems.
Variations of this model continuously produce new state-of-the-art results in different tasks.
The main difference between these variants lies in how the pretraining tasks are designed to take advantage of the latent information in the data.
Pretrained models such as BERT~\cite{devlin2018bert},  GPTs~\cite{radford2018improving,radford2019language,brown2020language},  ALBERT~\cite{lan2019albert}, ELECTRA~\cite{clark2020electra}, and BART~\cite{lewis2019bart} are all based on Transfomer but have different approaches to the pretraining tasks.
Hence, proposals of pretraining tasks are essential contributions to the development of pretrained models.

Transformer-based pretrained models all need effective pretraining tasks to learn latent patterns in the data.
The authors of BERT use two tasks, \textit{masked language modeling} and \textit{next sentence prediction} to train this model.
The idea of the \textit{masked language modeling} task is that when some words are masked, a good language model should be able to recover the original words.
The \textit{next sentence prediction} is the task that requires BERT to determine whether a sentence is the next one of another sentence.

GPT is a language model that is trained to recognize the next word of a set of given words, the same approach as N-gram language model.
Based on GPT, later versions of this model with a huge number of parameters are able to perform different tasks with very few training samples (few-shot learning) or even no training samples at all (zero-shot learning).
GPTs authors also introduce the concept of task conditioning which means with the same input, in different tasks the model must output differently. 
Language models with patterns learned from data can perform many impressive tasks.

In addition to the pretraining tasks of BERT and GPT, there are other proposals that help to improve the effectiveness or efficiency of the model.
ALBERT replaces BERT's \textit{next sentence prediction} task with \textit{sentence order prediction}. 
Instead of simply concluding whether two sentences are consecutive or not, the model needs to predict the order of two consecutive sentences.
ELECTRA's authors proposed \textit{replaced token detection} as an alternative to \textit{the masked language modeling} task. 
With a discriminator and a generator parallelly trained, the language model needs to find out which token is authentic, which is replaced.
BART is considered as a combination of BERT and GPT. This model has both prediction and generation capabilities. A series of pretraining tasks which is applied to BART are \textit{Token Masking}, \textit{Token Deletion}, \textit{Text Infilling}, \textit{Sentence Permutation}, and \textit{Document Rotation}.

Pretraining tasks are usually formed based on existing data structures.
The language modeling tasks are based on the consecutive and co-occurrence structure of words, sentences, and paragraphs.
Additionally, there exist many natural data structures that help pretrained models increase their performance.
Detecting specific structures contained in data is important to formulate the corresponding tasks.

\begin{figure}[htbp]
  \centering
  \includegraphics[width=\columnwidth]{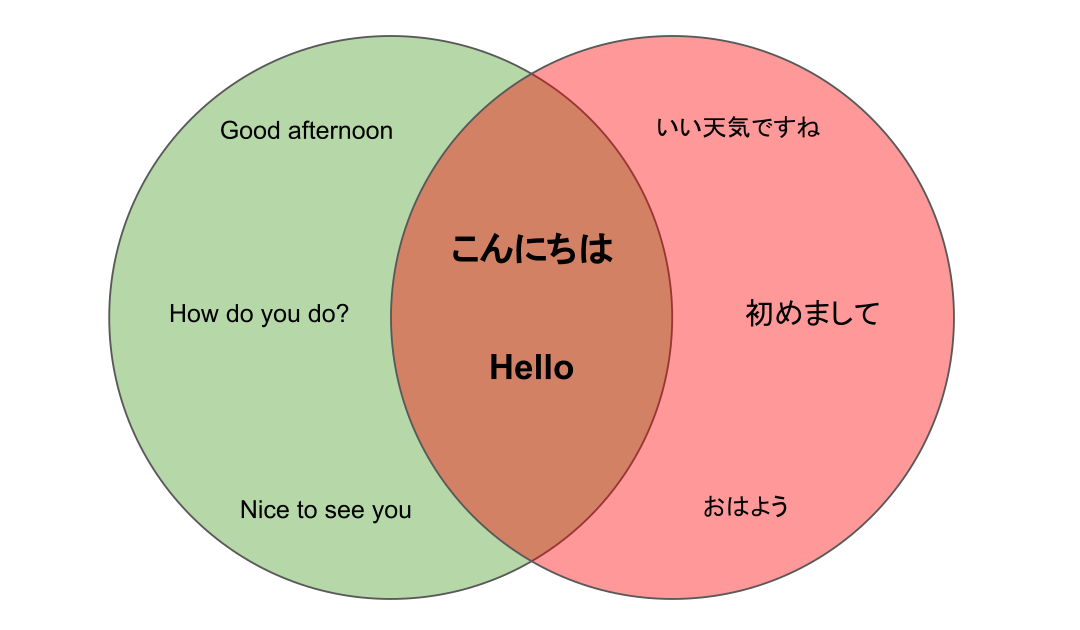}
  \caption{A single word may have multiple translations.}
  \label{fig:translation}
\end{figure}

A translation of a text gives us more information about its meaning than just a set of vocabulary translated into a new language.
A sentence in a language may contain many different semantics and depending on the context, the translation needs to be the most appropriate sentence in the target language with the same meaning.
For example, as in Figure~\ref{fig:translation} in Japanese, \begin{CJK*}{UTF8}{min}こんにちは\end{CJK*} can be a midday greeting or a formal way to say "hello". In consequence, in the morning context, this sentence needs to be translated as "hello" rather than "good afternoon".
Likewise, "I" in English can be translated in a multitude of ways in Japanese. 
Determining which is the correct translation must depend on the context of the sentence.

It is important to determine the correct context to correctly understand the meaning of a sentence when dealing with difficult documents such as the law.
A correct understanding of semantic will not depend on its language of expression.
Therefore, using the original version and the translation in parallel can help the model learn the semantic with better precision.

From such observations, we propose ParaLaw Nets, cross-lingual sentence-level pretraining models for legal document processing.
The idea is to force the model to learn the context dependence from bilingual sentences.
We conduct experiments on COLIEE-2021 data to verify the effectiveness of the method.
Our approach is superior to other multilingual approaches such as BERT Multilingual or XLM-RoBERTa.

\section{Related Work}
\label{section:related-works}
\subsection{Multilingual and Crosslingual Approaches}
The NLP resources are not uniform across languages.
The language with the most abundant resources is English. That of other languages is usually much less.
Therefore, resources developed in English will later be transferred to other languages.
The multilingual nature of resources is often understood as a translation into other languages from English.
The aspect of using semantic in multilingual to reduce ambiguity is also worth investigating.

The multilingual implementation was introduced for the first time when the authors of BERT~\cite{devlin2018bert} presented this kind of pretrained models. 
However, the article does not mention in detail how to build this variant of BERT.
Fortunately, on their Github~\footnote{https://github.com/google-research}, the authors state that this model is trained with the 100 largest Wikipedia languages.
Common languages are downsampled and less common languages are upsampled to ensure the patterns are learnable across different languages.

Lample et al. \cite {lample2019cross} proposed the idea of pretraining models using multiple languages. The authors use 3 tasks to train the model: \textit{Causal Language Modeling} (CLM), \textit{Masked Language Modeling} (MLM) and \textit{Translation Language Modeling} (TLM). Among them, TLM is a task that requires many languages, the model uses translation knowledge between languages to fill the missing words in the blanks. According to the authors, this task forces the model to learn the alignment between languages and leverage the context of one language when the context of the other language is not complete.

XLM-RoBERTa \cite{conneau2019unsupervised} is a pretrained model that uses multilingual advantages over 100 languages. 
This model with a huge amount of training data on these languages achieved state-of-the-art results on different tasks on the GLUE benchmark compared with cross-language baselines.
Through the article, the authors also prove the superiority of multilingual models compared with single-language models.

\subsection{Pretrained Models in Legal Domain}
In the NLP domains, legal document processing needs particular approaches.
The legal vocabulary is different from ordinary language, and law sentences often have a complex structure.
Pretrained methods in the legal domain have been proved to be competitive with other methods.
Most COLIEE-2020 approaches, including the best systems, use pretrained models \cite{rabelocoliee}.

The Task 1 winner, \textit{cyber} team uses a pretrained Transformer model in their implementation for vector representation~\cite{westermann2020coliee}.
To solve Task 2, \textit{JNLP} team \cite{nguyen2020jnlp} uses the pretrained model based on supporting information to find supporting paragraphs across the legal cases.
The Task 3 winner, \textit{LLNTU} team \cite{shao2020coliee} uses BERT to classify whether an article is relevant to a given legal question or not.
For Task 4, \textit{JNLP} team also pretrains a legal language model from the case law data to generate strong contextual embedding for the model before making predictions in the statute law.
With the limited data and narrow specializations, the pretrained models seem to be a competitive approach.

From observing that multilingual information can support to model the meaning of sentences, we propose Paralaw Nets, pretrained models using multilingual pairs of legal sentences.
In any linguistic task, especially in the legal domain, it is very important to understand correctly the meaning of a sentence in making predictions.
By forcing the model to learn the possibility of the semantic connection between the two sentences, we believe in having a strong pretrained model.

\section{Paralaw Nets}
\label{section:proposed-approach}

\subsection{Pretraining}
The general idea of this paper's approach is to utilize the hidden information that is aligned between two sentences in two different languages to train the model.
Different from the token-based approach of XLM-RoBERTa, we use sentence-level approach.
The semantic understanding ability of a model is judged on how well it predicts the logical order of sentences in different languages.
We propose two approaches called \textit{Next Foreign Sentence Prediction} (NFSP) and \textit{Neighbor Multilingual Sentence Prediction} (NMSP).

Not only considering multilingual as an additional version of the pretrained models in English, we believe that the translation information will help the model to have a better understanding of the sentence meaning.
When an idea is correctly understood by a model, this model can verify the expression of that idea in all languages.
In other words, semantic is expressed through, but not limited by, the language in which it is expressed.

In the NFSP task, the model needs to read two sentences in different languages and determine if their semantic belong to two consecutive sentences in a document.
To this end, the model needs to correctly understand the meaning of each sentence.
This is intended to reduce ambiguity in the expression of sentences in both languages.

For example, from original sentences as "The weather is nice. Shall we go out?" and their translations \begin{CJK*}{UTF8}{min}"いい天気ね。お出掛けしよ？"\end{CJK*}, we can create 2 positive samples in the training data as:

\begin{CJK*}{UTF8}{min}
\begin{itemize}
    \item The weather is nice. お出掛けしよ？

    \item いい天気ね。Shall we go out?
\end{itemize}
\end{CJK*}

The negative samples are pairs of a sentence in the original documents and the translation of another random sentence.

NMSP shares a similar approach with NFSP but the training data is generated with more cases.
In addition to bilingual pairs, we include pairs of same-language sentences in two languages.
If cross-lingual factor is not considered, NFSP has the same approach as BERT's NSP task.
BERT's NSP critics argue that the Transformer model can rely on co-appearance information to predict the labels of samples in this task.
That is, even if the model does not know whether two sentences are consecutive or not, it can still guess the label based on the proximity of the topic.
Dealing with that potential issue, our hypothesis is that if the model determines that sentence A follows sentence B, it must also know that sentence B comes before sentence A.
In addition, we assign one label to help the model learn that two sentences are not contiguous in a text.

NFSP is a binary classification problem, NMSP is a multi-label classification problem with labels corresponding to the case of random sampling, normal order, reverse order, and non-contiguous.

The data used to pretrain our Paralaw Nets is bilingual legal sentences.
Thanks to globalization, legal documents in other languages are often translated into English sentence by sentence.
This creates a great edge for ParaLaw Nets in terms of pretraining data.
The experimental models introduced in this paper are trained with Japanese-English bilingual legal data. 
However, this approach can be generalized to all language pairs or groups.

We pretrain models according to the tasks mentioned. The BERT multilingual base model is used as the base model for both NFSP and NMSP. The distilled versions use the configuration and architecture of BERTDistilled \cite{sanh2019distilbert}. All models are cased configurations. Data for pretraining NFSP contains 239,000 samples, data for pretraining NMSP contains 718,000 samples. The training process is stopped when the performance of the model does not increase on the validation set. Table~\ref{tab:pretraining_paramters} shows the parameters and the performances in pretraining the models.

\begin{table*}[]
\caption{Parameters and performances in pretraining the models}
\begin{tabular}{lcccc}
\hline
\textbf{Model} & \textbf{Max Length} & \textbf{Batch Size} & \textbf{Number of Batches} & \textbf{Validation Accuracy} \\
\hline
NFSP Base       & 512                 & 16                  & 24,000                     & 94.4\%                       \\
NFSP Distilled  & 512                 & 32                  & 34,000                     & 92.2\%                       \\
NMSP Base       & 512                 & 16                  & 320,000                    & 88.0\%                       \\
NMSP Distilled  & 512                 & 32                  & 496,000                    & 87.7\%                       \\
\hline
\end{tabular}

\label{tab:pretraining_paramters}
\end{table*}

\subsection{Finetuning}
Next, we finetune the models for the lawfulness classification problem in Task 5 of the COLIEE-2021. Given a statement as a legal question, the model needs to decide whether that statement is true or false.
Without the support of lexical-based retrieval systems, the model needs to really understand the meaning of the previously learned propositions, generalize them and apply that knowledge to the question. Table \ref{tab:task5_examples} shows examples of this task.

\begin{table*}[]
\caption{Examples for COLIEE-2021's Task 5}
\begin{tabular}{|l|r|}
\hline
\textbf{Sentence}                                                       & \textbf{Output} \\ \hline
No abuse of rights is permitted.                                        & Yes             \\ \hline
The age of majority is reached when a person has reached the age of 12. & No              \\ \hline
\end{tabular}

\label{tab:task5_examples}
\end{table*}

To strengthen the bilingual model, we use original and augmented data in both English and Japanese.
Negation is the main method used to create variations of original data.
The first negation rule that is matched will be used only once to avoid the negation of the negation.
With English negation rules, we reuse the rules proposed by Nguyen et al~\cite{jnlp_task4_coliee2019}.
Japanese negation rules are derived from basic Japanese syntax.
English and Japanese negation rules are shown in Tables \ref{tab:en_rules} and \ref{tab:ja_rules}.

\begin{table}[t]
  \caption{Rules applied for negation statement generation in English \cite{jnlp_task4_coliee2019}}
  \label{tab:en_rules}
  \begin{center}

  \begin{tabular}{|l|l|}
    \hline
    \textbf{Original Statement}& \textbf{Negation Statement Generation}\\
    \hline
    contains $not$&Remove $not$ from original statement\\
    contains $shall$&Replace $shall$ with $shall\ not$\\
    contains $should$&Replace $should$ with $should\ not$\\
    contains $may$&Replace $may$ with $may\ not$\\
    contains $must$&Replace $must$ with $must\ not$\\\hline
    contains $is$&Replace $is$ with $is\ not$\\
    contains $are$&Replace $are$ with $are\ not$\\
    contains $will\ be$&Replace $will\ be$ with $will\ not\ be$\\\hline
    contains $can$&Replace $can$ with $cannot$\\
    contains $cannot$&Replace $cannot$ with $can$\\
    contains $with$&Replace $with$ with $without$\\
    contains $without$&Replace $without$ with $with$\\\hline
    contains $A$&Replace $A$ with $No$\\
    contains $An$&Replace $An$ with $No$\\
  \hline
\end{tabular}
\end{center}
\end{table}

\begin{table*}[t]
  \caption{Rules applied for negation statement generation in Japanese}
  \label{tab:ja_rules}
  \begin{center}

  \begin{tabular}{|l|l|}
    \hline
    \textbf{Original Statement}& \textbf{Negation Statement Generation}\\
    \hline
    contains \begin{CJK*}{UTF8}{min}ません\end{CJK*}&Replace \begin{CJK*}{UTF8}{min}ません\end{CJK*} with \begin{CJK*}{UTF8}{min}ます\end{CJK*}\\
    contains \begin{CJK*}{UTF8}{min}できる\end{CJK*}&Replace \begin{CJK*}{UTF8}{min}できる\end{CJK*} with \begin{CJK*}{UTF8}{min}できない\end{CJK*}\\
    contains \begin{CJK*}{UTF8}{min}できない\end{CJK*}&Replace \begin{CJK*}{UTF8}{min}できない\end{CJK*} with \begin{CJK*}{UTF8}{min}できる\end{CJK*}\\
    contains \begin{CJK*}{UTF8}{min}した\end{CJK*}&Replace \begin{CJK*}{UTF8}{min}した\end{CJK*} with \begin{CJK*}{UTF8}{min}しなかった\end{CJK*}\\
    contains \begin{CJK*}{UTF8}{min}でない\end{CJK*}&Replace \begin{CJK*}{UTF8}{min}でない\end{CJK*} with \begin{CJK*}{UTF8}{min}である\end{CJK*}\\
    contains \begin{CJK*}{UTF8}{min}できた\end{CJK*}&Replace \begin{CJK*}{UTF8}{min}できた\end{CJK*} with \begin{CJK*}{UTF8}{min}できなかった\end{CJK*}\\
    contains \begin{CJK*}{UTF8}{min}させる\end{CJK*}&Replace \begin{CJK*}{UTF8}{min}させる\end{CJK*} with \begin{CJK*}{UTF8}{min}させない\end{CJK*}\\
    contains \begin{CJK*}{UTF8}{min}ている\end{CJK*}&Replace \begin{CJK*}{UTF8}{min}ている\end{CJK*} with \begin{CJK*}{UTF8}{min}ていない\end{CJK*}\\
    contains \begin{CJK*}{UTF8}{min}がない\end{CJK*}&Replace \begin{CJK*}{UTF8}{min}がない\end{CJK*} with \begin{CJK*}{UTF8}{min}がある\end{CJK*}\\
    contains \begin{CJK*}{UTF8}{min}ではない\end{CJK*}&Replace \begin{CJK*}{UTF8}{min}ではない\end{CJK*} with \begin{CJK*}{UTF8}{min}ではある\end{CJK*}\\
    contains \begin{CJK*}{UTF8}{min}ことがある\end{CJK*}&Replace \begin{CJK*}{UTF8}{min}ことがある\end{CJK*} with \begin{CJK*}{UTF8}{min}ことがない\end{CJK*}\\
    contains \begin{CJK*}{UTF8}{min}しなければならない\end{CJK*}&Replace \begin{CJK*}{UTF8}{min}しなければならない\end{CJK*} with \begin{CJK*}{UTF8}{min}してはいけません\end{CJK*}\\
    contains \begin{CJK*}{UTF8}{min}ならない\end{CJK*}&Replace \begin{CJK*}{UTF8}{min}ならない\end{CJK*} with \begin{CJK*}{UTF8}{min}なる\end{CJK*}\\
    
  \hline
\end{tabular}
\end{center}
\end{table*}

\section{Experiments}
\label{section:experiments}

\subsection{Experimental Setup}
We do experiments to choose the best models to generate predictions on the blind test set of COLIEE-2021's organizer.
Data in English includes all data provided by the organizer and a portion of the Japanese Civil Code.
In the Japanese Civil Code, statements that are represented as lists are removed because their elements are often lengthy and do not express a complete semantic. 
In addition, it is not a valid approach if we concatenate them without carefully considering the logical semantic of the whole statement.
For example, in natural language, \textit{and/or} conjunction in a sentence may differ from the logical meaning which the sentence expresses.
The process of filtering sentences is processed completely automatically based on the XML structure provided by the Japanese Law Translation website~\footnote{https://www.japaneselawtranslation.go.jp}.

We augment the data by negation rules as described in Section \ref{section:proposed-approach}.
All full sentences in the Japanese Civil Code are considered lawful and their negations are unlawful.
With the data provided by the organizers, the sentences already have labels, we create more data by creating negation of the content and reversing the labels.
Data after augmentation contains 7,000 sentences, we use 10\% for validation data, the rest is for training.

We experiment on the lawfulness classification problem with 6 different models including the original BERT multilingual base model from Google, XLM-RoBERTa, NFSP base, NFSP distilled, NMSP base and NMSP distilled.

\subsection{Experimental Results}
Training
the models, we observed interesting phenomena when training the models using Japanese data.
If we use all the data is augmented with the rules in Table \ref{tab:ja_rules}, all models cannot converge.
To solve this problem, we use a simple curriculum learning strategy for Japanese data. We train 3 epochs using augmented data by the first three negation rules before training with the whole dataset.
With only the English data, we did not encounter this problem.
We believe that this is an indication of the more challenges in understanding Japanese versus understanding English for the cross-lingual models trained with our approach.
Looking at Table \ref{tab:en_rules} and Table \ref{tab:ja_rules}, it can be seen that the English negations are related to the word "not", the negations of Japanese are more diverse and complex. Therefore the model needs more skills to distinguish negation.

\begin{figure}[]
    \centering
    \includegraphics[width=.4\textwidth]{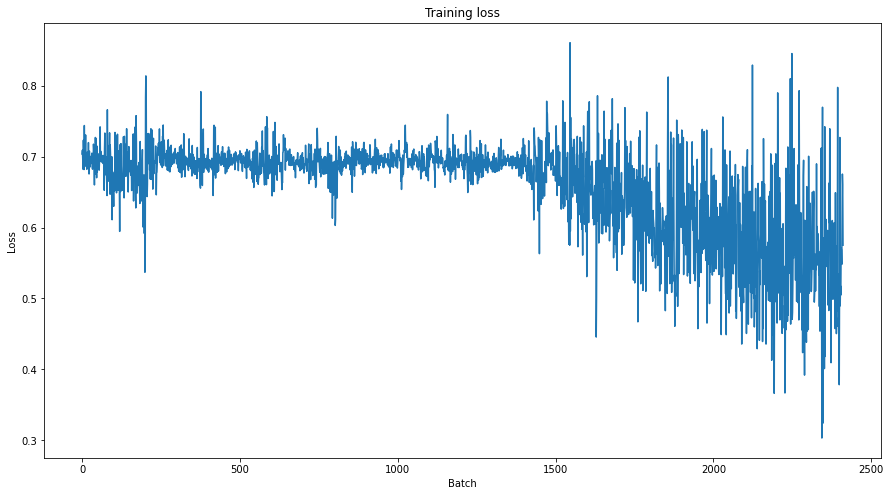}
    \caption{Loss fluctuation of BERT Multilingual.}
    \label{fig:multilingual_loss}
    
\end{figure}

\begin{figure}[]
    \label{fig:xlm_loss}
    \centering
    \includegraphics[width=.4\textwidth]{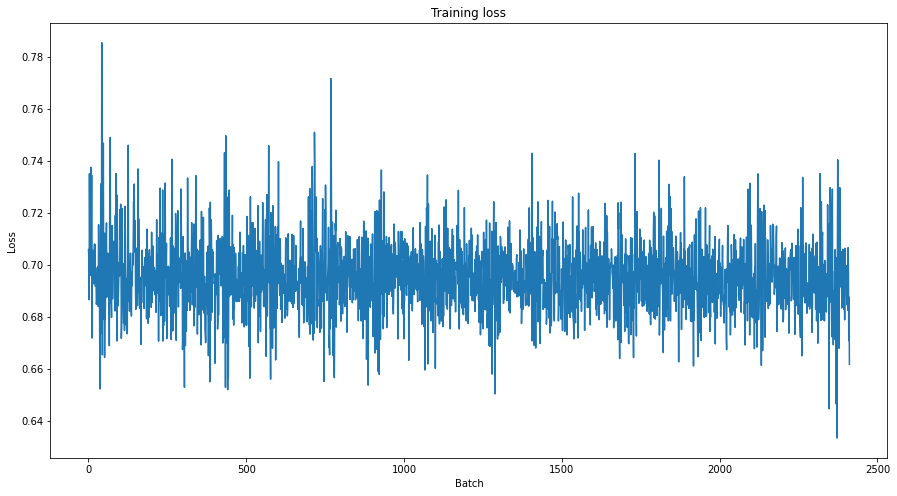}
    \caption{Loss fluctuation of XLM-RoBERTa.}
    
\end{figure}

\begin{figure}[]
    \label{fig:nmsp_base_loss}
    \centering
    \includegraphics[width=.4\textwidth]{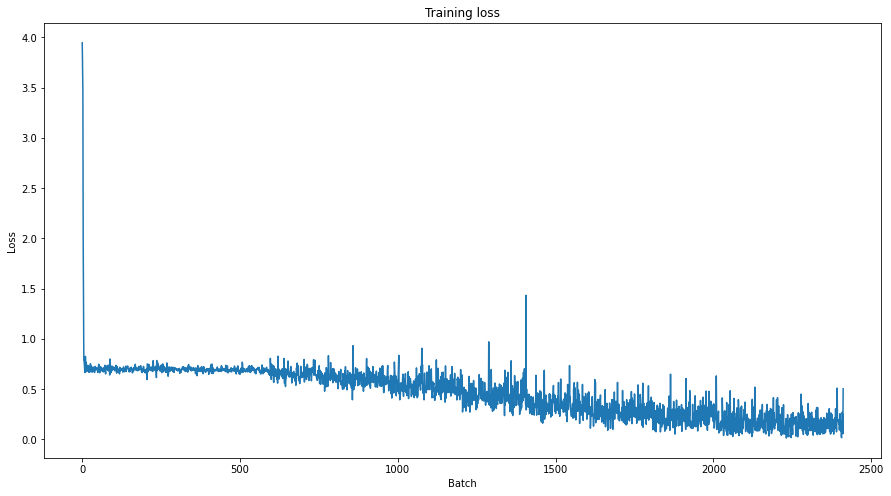}
    \caption{Loss fluctuation of NMSP Base.}
    
\end{figure}

\begin{figure}[]
    \label{fig:nmsp_distilled_loss}
    \centering
    \includegraphics[width=.4\textwidth]{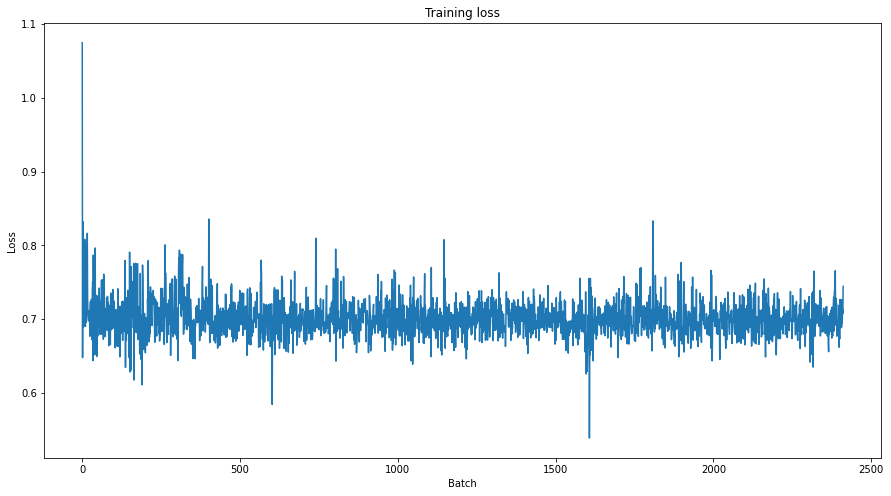}
    \caption{Loss fluctuation of NMSP Distilled.}
    
\end{figure}

\begin{figure}[]
    \label{fig:nfsp_base_loss}
    \centering
    \includegraphics[width=.4\textwidth]{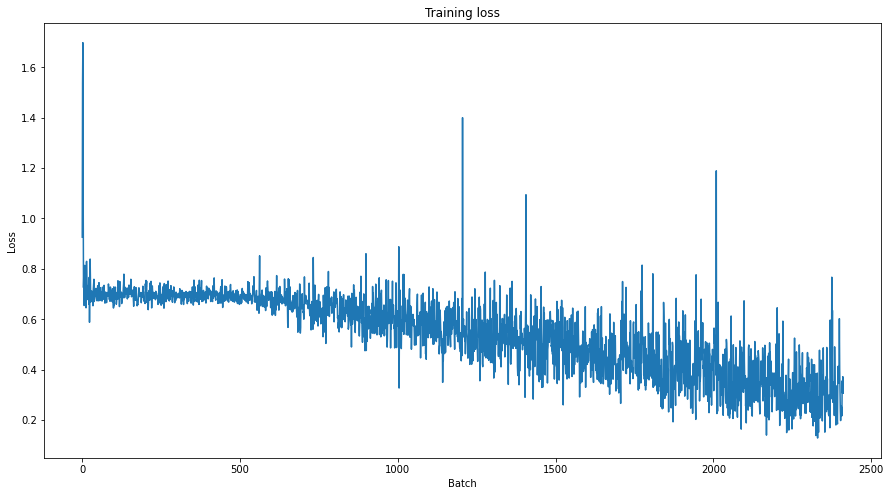}
    \caption{Loss fluctuation of NFSP Base.}
    
\end{figure}

\begin{figure}[]
    
    \centering
    \includegraphics[width=.4\textwidth]{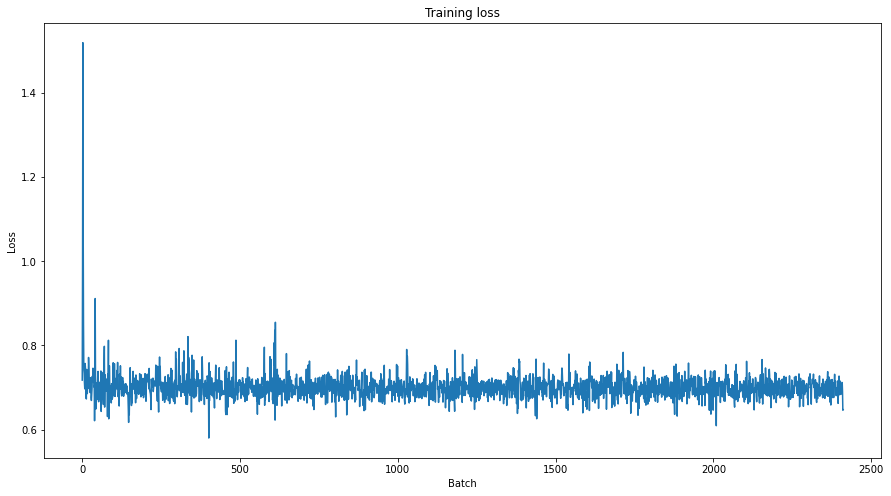}
    \caption{Loss fluctuation of NFSP Distilled.}
    \label{fig:nfsp_distilled_loss}
\end{figure}

Table \ref{tab:finetune_result_validation} shows the performance of the models on the validation set. The distilled models and XLM-RoBERTa completely fail to learn in this task. Figures \ref{fig:multilingual_loss}-\ref{fig:nfsp_distilled_loss} plot the loss fluctuation of these models. The loss values fluctuate around 0.7 and do not decrease. The original BERT Multilingual model passes the threshold of 0.7. Although this model has a huge variant loss value after that, its accuracy is better than XLM-RoBERTa and the distilled models. NFSP Base and NMSP Base have loss reduced to below 0.7 and loss variation is much more stable.

From the experimental results, we choose three candidates for final runs: NFSP Base, NMSP Base and Original BERT Multilingual.

\begin{table}[]
\caption{Performance of models on validation set}
\label{tab:finetune_result_validation}
\begin{tabular}{lr}

\toprule
\textbf{Model}    & \textbf{Accuracy} \\
\toprule
NFSP Base          & 71.0\%                       \\
NFSP Distilled     & 51.1\%                       \\
\midrule
NMSP Base          & 79.5\%                       \\
NMSP Distilled     & 48.9\%                       \\
\midrule
XLM-RoBERTa       & 51.1\%                       \\
BERT Multilingual & 64.1\%                      \\
\bottomrule
\end{tabular}
\end{table}

\subsection{Final Runs Result}
We run predictions on the English blind test set provided by the COLIEE-2021's organizer. Among our 3 models, NFSP Base has the best result, next is NMSP Base, and BERT Multilingual has the lowest result. We were also surprised that NFSP Base outperformed NMSP Base and stayed first on the leaderboard.
This may indicate that the test set distribution is somewhat biased against the latent features that the NFSP learned, which is not present in our validation set. However, the test set results support the notion that pretraining with cross-lingual information by our approach helps the model learn more accurately on finetuned tasks.

\begin{table*}[]
\caption{Result of final runs on the test set, the underlined lines refer to our models}

\begin{tabular}{|l|l|l|r|}
\hline
\textbf{Team} & \textbf{Run ID}                     & \textbf{Correct} & \textbf{Accuracy}     \\ \hline
              & BaseLine                            & No 43/All 81     & 0.5309                \\ \hline
{\ul JNLP}    & {\ul JNLP.NFSP}                     & {\ul 49}         & {\ul \textbf{0.6049}} \\ \hline
UA            & UA\_parser                          & 46               & 0.5679                \\ \hline
{\ul JNLP}    & {\ul JNLP.NMSP}                     & {\ul 45}         & {\ul 0.5556}          \\ \hline
UA            & UA\_dl                              & 45               & 0.5556                \\ \hline
TR            & TRDistillRoberta                    & 44               & 0.5432                \\ \hline
KIS           & KIS\_2                              & 41               & 0.5062                \\ \hline
KIS           & KIS\_3                              & 41               & 0.5062                \\ \hline
UA            & UA\_elmo                            & 40               & 0.4938                \\ \hline
{\ul JNLP}    & {\ul JNLP.BERT\_Multilingual} & {\ul 38}         & {\ul 0.4691}          \\ \hline
KIS           & KIS\_1                              & 35               & 0.4321                \\ \hline
TR            & TRGPT3Ada                           & 35               & 0.4321                \\ \hline
TR            & TRGPT3Davinci                       & 35               & 0.4321                \\ \hline
\end{tabular}
\label{tab:final_runs}

\end{table*}

\section{Conclusions}
\label{section:conclusion}
This paper proposes an approach using sentence-level cross-lingual information to pretrain transformer models before finetuning on the specific task. Taking advantage of cross-lingual resources in legal documents, we introduce NFSP and NMSP models which have impressive performance in our experiments as well as in COLIEE-2021's blind test. The idea of this study is applicable to problems with aligned translation data as legal text processing.
\noindent
\begin{acks}
This work was supported by JSPS KAKENHI Grant Numbers JP17H06103 and
JP20K20406.
\end{acks}

\bibliographystyle{ACM-Reference-Format}
\bibliography{references}


\begin{thebibliography}{16}


\ifx \showCODEN    \undefined \def \showCODEN     #1{\unskip}     \fi
\ifx \showDOI      \undefined \def \showDOI       #1{#1}\fi
\ifx \showISBNx    \undefined \def \showISBNx     #1{\unskip}     \fi
\ifx \showISBNxiii \undefined \def \showISBNxiii  #1{\unskip}     \fi
\ifx \showISSN     \undefined \def \showISSN      #1{\unskip}     \fi
\ifx \showLCCN     \undefined \def \showLCCN      #1{\unskip}     \fi
\ifx \shownote     \undefined \def \shownote      #1{#1}          \fi
\ifx \showarticletitle \undefined \def \showarticletitle #1{#1}   \fi
\ifx \showURL      \undefined \def \showURL       {\relax}        \fi
\providecommand\bibfield[2]{#2}
\providecommand\bibinfo[2]{#2}
\providecommand\natexlab[1]{#1}
\providecommand\showeprint[2][]{arXiv:#2}

\bibitem[\protect\citeauthoryear{Brown, Mann, Ryder, Subbiah, Kaplan, Dhariwal,
  Neelakantan, Shyam, Sastry, Askell, et~al\mbox{.}}{Brown
  et~al\mbox{.}}{2020}]%
        {brown2020language}
\bibfield{author}{\bibinfo{person}{Tom~B Brown}, \bibinfo{person}{Benjamin
  Mann}, \bibinfo{person}{Nick Ryder}, \bibinfo{person}{Melanie Subbiah},
  \bibinfo{person}{Jared Kaplan}, \bibinfo{person}{Prafulla Dhariwal},
  \bibinfo{person}{Arvind Neelakantan}, \bibinfo{person}{Pranav Shyam},
  \bibinfo{person}{Girish Sastry}, \bibinfo{person}{Amanda Askell},
  {et~al\mbox{.}}} \bibinfo{year}{2020}\natexlab{}.
\newblock \showarticletitle{Language models are few-shot learners}.
\newblock \bibinfo{journal}{\emph{arXiv preprint arXiv:2005.14165}}
  (\bibinfo{year}{2020}).
\newblock


\bibitem[\protect\citeauthoryear{Clark, Luong, Le, and Manning}{Clark
  et~al\mbox{.}}{2020}]%
        {clark2020electra}
\bibfield{author}{\bibinfo{person}{Kevin Clark}, \bibinfo{person}{Minh-Thang
  Luong}, \bibinfo{person}{Quoc~V Le}, {and} \bibinfo{person}{Christopher~D
  Manning}.} \bibinfo{year}{2020}\natexlab{}.
\newblock \showarticletitle{Electra: Pre-training text encoders as
  discriminators rather than generators}.
\newblock \bibinfo{journal}{\emph{arXiv preprint arXiv:2003.10555}}
  (\bibinfo{year}{2020}).
\newblock


\bibitem[\protect\citeauthoryear{Conneau, Khandelwal, Goyal, Chaudhary, Wenzek,
  Guzm{\'a}n, Grave, Ott, Zettlemoyer, and Stoyanov}{Conneau
  et~al\mbox{.}}{2019}]%
        {conneau2019unsupervised}
\bibfield{author}{\bibinfo{person}{Alexis Conneau}, \bibinfo{person}{Kartikay
  Khandelwal}, \bibinfo{person}{Naman Goyal}, \bibinfo{person}{Vishrav
  Chaudhary}, \bibinfo{person}{Guillaume Wenzek}, \bibinfo{person}{Francisco
  Guzm{\'a}n}, \bibinfo{person}{Edouard Grave}, \bibinfo{person}{Myle Ott},
  \bibinfo{person}{Luke Zettlemoyer}, {and} \bibinfo{person}{Veselin
  Stoyanov}.} \bibinfo{year}{2019}\natexlab{}.
\newblock \showarticletitle{Unsupervised cross-lingual representation learning
  at scale}.
\newblock \bibinfo{journal}{\emph{arXiv preprint arXiv:1911.02116}}
  (\bibinfo{year}{2019}).
\newblock


\bibitem[\protect\citeauthoryear{Devlin, Chang, Lee, and Toutanova}{Devlin
  et~al\mbox{.}}{2018}]%
        {devlin2018bert}
\bibfield{author}{\bibinfo{person}{Jacob Devlin}, \bibinfo{person}{Ming-Wei
  Chang}, \bibinfo{person}{Kenton Lee}, {and} \bibinfo{person}{Kristina
  Toutanova}.} \bibinfo{year}{2018}\natexlab{}.
\newblock \showarticletitle{Bert: Pre-training of deep bidirectional
  transformers for language understanding}.
\newblock \bibinfo{journal}{\emph{arXiv preprint arXiv:1810.04805}}
  (\bibinfo{year}{2018}).
\newblock


\bibitem[\protect\citeauthoryear{Hannes, Jaromir, and Karim}{Hannes
  et~al\mbox{.}}{2020}]%
        {westermann2020coliee}
\bibfield{author}{\bibinfo{person}{Westermann Hannes}, \bibinfo{person}{Savelka
  Jaromir}, {and} \bibinfo{person}{Benyekhlef Karim}.}
  \bibinfo{year}{2020}\natexlab{}.
\newblock \showarticletitle{Paragraph Similarity Scoring and Fine-Tuned BERT
  for Legal Information Retrieval and Entailment}.
\newblock \bibinfo{journal}{\emph{COLIEE 2020}} (\bibinfo{year}{2020}).
\newblock


\bibitem[\protect\citeauthoryear{Hsuan-Lei, Yi-Chia, and Sieh-Chuen}{Hsuan-Lei
  et~al\mbox{.}}{2020}]%
        {shao2020coliee}
\bibfield{author}{\bibinfo{person}{Shao Hsuan-Lei}, \bibinfo{person}{Chen
  Yi-Chia}, {and} \bibinfo{person}{Huang Sieh-Chuen}.}
  \bibinfo{year}{2020}\natexlab{}.
\newblock \showarticletitle{BERT-based Ensemble Model for The Statute Law
  Retrieval and Legal Information Entailment}.
\newblock \bibinfo{journal}{\emph{COLIEE 2020}} (\bibinfo{year}{2020}).
\newblock


\bibitem[\protect\citeauthoryear{Lample and Conneau}{Lample and
  Conneau}{2019}]%
        {lample2019cross}
\bibfield{author}{\bibinfo{person}{Guillaume Lample} {and}
  \bibinfo{person}{Alexis Conneau}.} \bibinfo{year}{2019}\natexlab{}.
\newblock \showarticletitle{Cross-lingual language model pretraining}.
\newblock \bibinfo{journal}{\emph{arXiv preprint arXiv:1901.07291}}
  (\bibinfo{year}{2019}).
\newblock


\bibitem[\protect\citeauthoryear{Lan, Chen, Goodman, Gimpel, Sharma, and
  Soricut}{Lan et~al\mbox{.}}{2019}]%
        {lan2019albert}
\bibfield{author}{\bibinfo{person}{Zhenzhong Lan}, \bibinfo{person}{Mingda
  Chen}, \bibinfo{person}{Sebastian Goodman}, \bibinfo{person}{Kevin Gimpel},
  \bibinfo{person}{Piyush Sharma}, {and} \bibinfo{person}{Radu Soricut}.}
  \bibinfo{year}{2019}\natexlab{}.
\newblock \showarticletitle{Albert: A lite bert for self-supervised learning of
  language representations}.
\newblock \bibinfo{journal}{\emph{arXiv preprint arXiv:1909.11942}}
  (\bibinfo{year}{2019}).
\newblock


\bibitem[\protect\citeauthoryear{Lewis, Liu, Goyal, Ghazvininejad, Mohamed,
  Levy, Stoyanov, and Zettlemoyer}{Lewis et~al\mbox{.}}{2019}]%
        {lewis2019bart}
\bibfield{author}{\bibinfo{person}{Mike Lewis}, \bibinfo{person}{Yinhan Liu},
  \bibinfo{person}{Naman Goyal}, \bibinfo{person}{Marjan Ghazvininejad},
  \bibinfo{person}{Abdelrahman Mohamed}, \bibinfo{person}{Omer Levy},
  \bibinfo{person}{Ves Stoyanov}, {and} \bibinfo{person}{Luke Zettlemoyer}.}
  \bibinfo{year}{2019}\natexlab{}.
\newblock \showarticletitle{Bart: Denoising sequence-to-sequence pre-training
  for natural language generation, translation, and comprehension}.
\newblock \bibinfo{journal}{\emph{arXiv preprint arXiv:1910.13461}}
  (\bibinfo{year}{2019}).
\newblock


\bibitem[\protect\citeauthoryear{Nguyen, Tran, and Nguyen}{Nguyen
  et~al\mbox{.}}{2019}]%
        {jnlp_task4_coliee2019}
\bibfield{author}{\bibinfo{person}{HT Nguyen}, \bibinfo{person}{V Tran}, {and}
  \bibinfo{person}{LM Nguyen}.} \bibinfo{year}{2019}\natexlab{}.
\newblock \showarticletitle{A deep learning approach for statute law entailment
  task in COLIEE-2019}.
\newblock \bibinfo{journal}{\emph{Proceedings of the 6th Competition on Legal
  Information Extraction/Entailment. COLIEE}} (\bibinfo{year}{2019}).
\newblock


\bibitem[\protect\citeauthoryear{Nguyen, Vuong, Nguyen, Dang, Bui, Vu, Nguyen,
  Tran, Satoh, and Nguyen}{Nguyen et~al\mbox{.}}{2020}]%
        {nguyen2020jnlp}
\bibfield{author}{\bibinfo{person}{Ha-Thanh Nguyen},
  \bibinfo{person}{Hai-Yen~Thi Vuong}, \bibinfo{person}{Phuong~Minh Nguyen},
  \bibinfo{person}{Binh~Tran Dang}, \bibinfo{person}{Quan~Minh Bui},
  \bibinfo{person}{Sinh~Trong Vu}, \bibinfo{person}{Chau~Minh Nguyen},
  \bibinfo{person}{Vu Tran}, \bibinfo{person}{Ken Satoh}, {and}
  \bibinfo{person}{Minh~Le Nguyen}.} \bibinfo{year}{2020}\natexlab{}.
\newblock \showarticletitle{JNLP Team: Deep Learning for Legal Processing in
  COLIEE 2020}.
\newblock \bibinfo{journal}{\emph{arXiv preprint arXiv:2011.08071}}
  (\bibinfo{year}{2020}).
\newblock


\bibitem[\protect\citeauthoryear{Rabelo, Kim, Goebel, Yoshioka, Kano, and
  Satoh}{Rabelo et~al\mbox{.}}{[n.d.]}]%
        {rabelocoliee}
\bibfield{author}{\bibinfo{person}{Juliano Rabelo}, \bibinfo{person}{Mi-Young
  Kim}, \bibinfo{person}{Randy Goebel}, \bibinfo{person}{Masaharu Yoshioka},
  \bibinfo{person}{Yoshinobu Kano}, {and} \bibinfo{person}{Ken Satoh}.}
  \bibinfo{year}{[n.d.]}\natexlab{}.
\newblock \showarticletitle{COLIEE 2020: Methods for Legal Document Retrieval
  and Entailment}.
\newblock  (\bibinfo{year}{[n.\,d.]}).
\newblock


\bibitem[\protect\citeauthoryear{Radford, Narasimhan, Salimans, and
  Sutskever}{Radford et~al\mbox{.}}{2018}]%
        {radford2018improving}
\bibfield{author}{\bibinfo{person}{Alec Radford}, \bibinfo{person}{Karthik
  Narasimhan}, \bibinfo{person}{Tim Salimans}, {and} \bibinfo{person}{Ilya
  Sutskever}.} \bibinfo{year}{2018}\natexlab{}.
\newblock \showarticletitle{Improving language understanding by generative
  pre-training}.
\newblock  (\bibinfo{year}{2018}).
\newblock


\bibitem[\protect\citeauthoryear{Radford, Wu, Child, Luan, Amodei, and
  Sutskever}{Radford et~al\mbox{.}}{2019}]%
        {radford2019language}
\bibfield{author}{\bibinfo{person}{Alec Radford}, \bibinfo{person}{Jeffrey Wu},
  \bibinfo{person}{Rewon Child}, \bibinfo{person}{David Luan},
  \bibinfo{person}{Dario Amodei}, {and} \bibinfo{person}{Ilya Sutskever}.}
  \bibinfo{year}{2019}\natexlab{}.
\newblock \showarticletitle{Language models are unsupervised multitask
  learners}.
\newblock \bibinfo{journal}{\emph{OpenAI blog}} \bibinfo{volume}{1},
  \bibinfo{number}{8} (\bibinfo{year}{2019}), \bibinfo{pages}{9}.
\newblock


\bibitem[\protect\citeauthoryear{Sanh, Debut, Chaumond, and Wolf}{Sanh
  et~al\mbox{.}}{2019}]%
        {sanh2019distilbert}
\bibfield{author}{\bibinfo{person}{Victor Sanh}, \bibinfo{person}{Lysandre
  Debut}, \bibinfo{person}{Julien Chaumond}, {and} \bibinfo{person}{Thomas
  Wolf}.} \bibinfo{year}{2019}\natexlab{}.
\newblock \showarticletitle{DistilBERT, a distilled version of BERT: smaller,
  faster, cheaper and lighter}.
\newblock \bibinfo{journal}{\emph{arXiv preprint arXiv:1910.01108}}
  (\bibinfo{year}{2019}).
\newblock


\bibitem[\protect\citeauthoryear{Vaswani, Shazeer, Parmar, Uszkoreit, Jones,
  Gomez, Kaiser, and Polosukhin}{Vaswani et~al\mbox{.}}{2017}]%
        {vaswani2017attention}
\bibfield{author}{\bibinfo{person}{Ashish Vaswani}, \bibinfo{person}{Noam
  Shazeer}, \bibinfo{person}{Niki Parmar}, \bibinfo{person}{Jakob Uszkoreit},
  \bibinfo{person}{Llion Jones}, \bibinfo{person}{Aidan~N Gomez},
  \bibinfo{person}{Lukasz Kaiser}, {and} \bibinfo{person}{Illia Polosukhin}.}
  \bibinfo{year}{2017}\natexlab{}.
\newblock \showarticletitle{Attention is all you need}.
\newblock \bibinfo{journal}{\emph{arXiv preprint arXiv:1706.03762}}
  (\bibinfo{year}{2017}).
\newblock


\end{thebibliography}

\appendix

\end{document}